%% file: main.tex
\documentclass[10pt,conference]{IEEEtran}
\IEEEoverridecommandlockouts
\usepackage{cite}
\usepackage{hyperref}
\usepackage{amsmath,amssymb,amsfonts}
\usepackage{balance}
\usepackage{algorithmic}
\usepackage{graphicx}
\usepackage{textcomp}
\usepackage{xcolor}
\def\BibTeX{{\rm B\kern-.05em{\sc i\kern-.025em b}\kern-.08em
    T\kern-.1667em\lower.7ex\hbox{E}\kern-.125emX}}
\usepackage{booktabs}
\usepackage{multirow}
\usepackage{caption}
\usepackage[caption=false]{subfig}
\usepackage{listings}
\usepackage{float}
\input{definition}

\begin{document}

\title{Estimating Predictive Uncertainty Under Program Data Distribution Shift}

\author{
\IEEEauthorblockN{Yufei Li}
\IEEEauthorblockA{
\textit{University of Texas at Dallas}\\
Dallas, USA \\
yxl190090@utdallas.edu}
\and
\IEEEauthorblockN{Simin Chen}
\IEEEauthorblockA{
\textit{University of Texas at Dallas}\\
Dallas, USA \\
simin.chen@utdallas.edu}
\and 
\IEEEauthorblockN{Wei Yang}
\IEEEauthorblockA{
\textit{University of Texas at Dallas}\\
Dallas, USA \\
wei.yang@utdallas.edu}

}

\maketitle

\input{Abstract}

\input{Introduction}

\input{Background}

\input{Distribution}

\input{Metric}

\input{Threats}
\input{Related}
\input{Future}
\input{Conclusion}


\balance
\bibliographystyle{IEEEtran}
\bibliography{IEEEabrv,ref}


\end{document}

%% file: definition.tex
\usepackage{xspace}    

\newcommand{\eg}{{\it e.g.,}\xspace}

%% file: Abstract.tex
\begin{abstract}

Deep learning (DL) techniques have achieved great success in predictive accuracy in a variety of tasks, but deep neural networks (DNNs) are shown to produce highly overconfident scores for even abnormal samples. Well-defined uncertainty indicates whether a model's output should (or should not) be trusted and thus becomes critical in real-world scenarios which typically involves shifted input distributions due to many factors. Existing uncertainty approaches assume that testing samples from a different data distribution would induce unreliable model predictions thus have higher uncertainty scores.
They quantify model uncertainty by calibrating DL model's confidence of a given input and  evaluate the effectiveness in computer vision (CV) and natural language processing (NLP)-related tasks. However, their methodologies' reliability may be compromised under programming tasks due to difference in data representations and  shift patterns. In this paper, we first define three different types of distribution shift in program data and build a large-scale shifted Java dataset. We implement two common programming language tasks on our dataset to study the effect of each distribution shift on DL model performance. We also propose a large-scale benchmark of existing state-of-the-art predictive uncertainty on programming tasks and investigate their effectiveness under data distribution shift. Experiments\footnote{Our implementation is available at \url{https://github.com/GAET-embedding/Uncertainty_Study.git}} show that program distribution shift does degrade the DL model performance to varying degrees and that existing uncertainty methods all present certain limitations in quantifying uncertainty on  program dataset.


\end{abstract}

%% file: Introduction.tex
\section{Introduction}

Recent success of deep learning (DL) \cite{lecun2015deep} application in a wide range of domains such as computer vision (CV), natural language processing (NLP) has attracted huge attention from researchers \cite{gal2016dropout}. Due to the superiority of DL techniques, they have also been broadly applied in nowadays software engineering (SE)-related tasks including autonomous driving testing, malware detection and programming language tasks. With the implementation of deep neural networks (DNNs), one can leverage the well-trained model to make predictions on the test dataset. The effectiveness of DL techniques is based on an important assumption that the test dataset is independently and identically distributed (i.i.d.) with the training dataset \cite{ovadia2019can}.

However, in practical scenarios once a model is deployed, the distribution over observed data may shift due to many factors including natural evolution \cite{gama2014survey}, noises \cite{carlini2017towards} and artificial adversarial attack \cite{wang2019adversarial}, and may eventually become much different from the original distribution. 
A typical example of distribution shift in programming language is software evolution \cite{mens2008introduction} which leads to various forms of code distribution shift, \eg programming language gets updated to a more recent library version, or the same project is taken over by another programmer with different writing habit.

Intuitively, test inputs from a shifted distribution can reduce DL model performance, but it is also critical to learn the specific impact that different types of data distribution shift has on DL models in terms of decision making. For example, if we know that the distribution shift caused by programming language update has little effect on program-analysis model performance, we do not need to retrain a new one. Retraining a DL model on a shifted dataset needs labelling on the dataset which requires large effort. And we need to strike a balance between the retraining cost and model performance degradation. Studying the impact of distribution shift on a model can facilitate us to understand when and how to adapt the model to the shifted dataset.




Moreover, when the testing distribution differs from the training distribution, DL models, though exhibit poor performance, still tend to assume their prediction is accurate and becomes untrustworthy \cite{amodei2016concrete}. Existing work \cite{lakshminarayanan2016simple, lee2018simple, ren2019likelihood, hendrycks2016baseline, hendrycks2020many, hendrycks2020pretrained, maddox2019simple} designs calibrated predictive uncertainty to evaluate the reliability of model's prediction of a given input. They assume abnormal samples such as out-of-distribution (OOD) and adversarial inputs are more likely to induce unreliable DL model predictions and thus have higher uncertainty scores. These uncertainty methods are widely used as an input validation mechanism to filter out uncertain inputs and satisfy accuracy requirement of (safety or security-critical) systems \cite{barrantes2003randomized}. 

\input{Figure/architecture}

However, to our knowledge all the existing state-of-the-art work evaluates the effectiveness of their uncertainty methods on CV and NLP tasks. Their conclusions may not be adaptive to programming language hypothesis due to the domain gap. 
The speciality of programming language task compared to CV and NLP in terms of distribution shift mainly lies in two aspects. First, their representations are different. Image models handle input pixel matrices, and the distribution shift over pixel matrix is generalizable statistical models such as Gaussian noise or linear matrix transformations such as image rotation, scaling, shear, etc. \cite{tian2018deeptest}; while program-analysis models, though may use the same DL architecture as NLP models such recurrent neural network (RNN), represent code snippets with a structured nature of their syntax such as abstract syntax tree (AST) rather than a linear sequence of tokens \cite{alon2019code2vec}. Therefore, the distribution shift on both NLP and programming data is not standard  generalizable manipulations as on CV data. Moreover, the distribution shift on code snippets contains additional structural relationships (\eg logic operations, function calls) compared to natural language.   
Second, their distribution shift sources are different. Shifted image datasets can be another dataset with total different discipline (\eg from pet image dataset to MNIST). Shifted natural language dataset can also be a text for another topic with different composition of words.
OOD datasets in both CV and NLP can be chosen from a different topic and present totally different meanings, while code data distribution shift has to follow the programming language grammar constraints and are generally milder such as language versions update, projects content change or programmer change.

To investigate the effectiveness of existing predictive uncertainty on programming language applications under data distribution shift, we define three different types of program data distribution shift based on real-world software development scenarios. Then we conduct a comprehensive study of the impact that each type of distribution shift has on DL model performance as well as on existing predictive uncertainty methods. We also analyze the advantages and limitations of each uncertainty method based on the evaluation results under program distribution shift for future study.

\noindent\textbf{Contributions.} The main contributions of this work are:

\begin{itemize}
    \item We build three shifted programming distribution datasets based on the real-world scenarios for future uncertainty study in software engineering.
    
    \item We  systematically study the effect that data distribution has on the DL model performance in two programming language classification tasks.
    
    \item We conduct a comprehensive study of the effectiveness of existing uncertainty methods on programming language tasks and the impact of data distribution on these uncertainty estimates.
    
    \item We summarize and analyze the limitations of existing uncertainty estimates on program tasks and datasets for future methodology refinement.
    
\end{itemize}



\noindent\textbf{Paper Organization.} The reminder of this paper is organized as follows. Section~\ref{sec:background} introduces necessary preliminaries. Section~\ref{sec:shift_effect} conducts study of effect of distribution shift on model performance. Section~\ref{sec:uncertainty_eval} evaluates the effectiveness of 5 popular uncertainty methods. Section~\ref{sec:threats} presents possible threats to validity of this article. Section~\ref{sec:related} introduces some related work we follow in this paper. Section~\ref{sec:future} proposes some future work. Section~\ref{sec:conclusion} summarizes our conclusion.

%% file: Figure/architecture.tex
\begin{figure}[htb]
    \centering
    \includegraphics[width=0.5\textwidth]{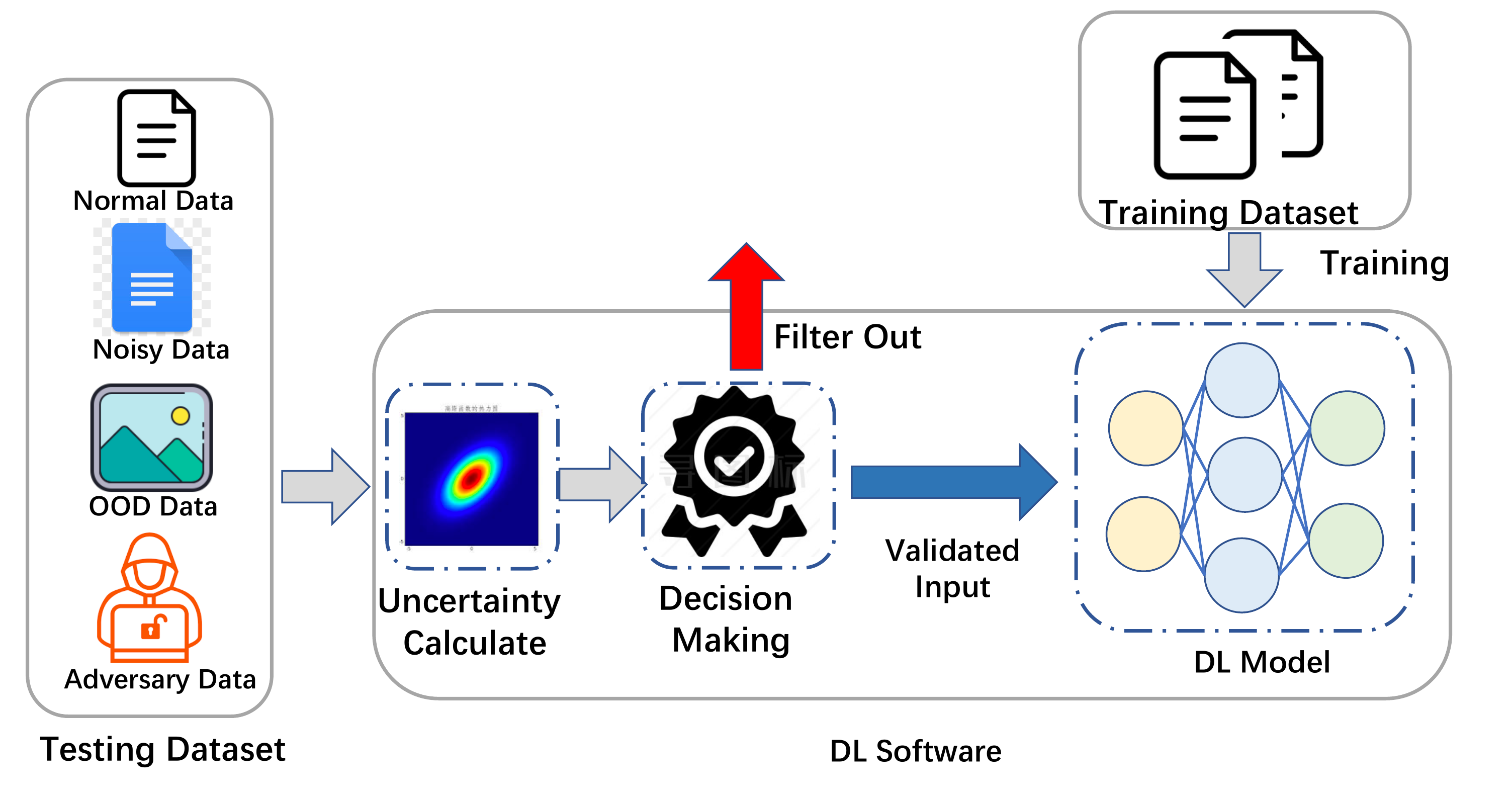}
    \caption{The architecture of input validation on DL software}
    \label{fig:arch}
\end{figure}

%% file: Background.tex
 \section{Preliminaries}
\label{sec:background}
DL models are used in DL software mainly for decision-making, which concerns the model training and input predication phases. Below we introduce the preliminaries of the two phases as
well as the related uncertainty background in DL applications.

\subsection{Uncertainty in Deep Learning}
With the deployment of DL systems in real-life settings, it becomes more critical to understand what a model does not know or cannot handle. 
DL systems can learn powerful representations but these mappings are often taken blindly and assumed to be accurate, which is not always the case. 
For example, In May 2016, there was the first fatality from an assisted driving system, caused by the perception system confusing the white side of a trailer for the bright sky.
This disaster was caused by the ignorance of the model's capability for recognizing white background.
Building reliable and truthful DL systems requires knowing the confidence behind the model's predictive probabilities. In other words, it requires digging deeply into uncertainty measurement.

The uncertainty in DL models arises from two sources, the \textit{aleatoric uncertainty} and the \textit{epistemic uncertainty} \cite{xiao2019quantifying}.
The former one arises from the noise in the observed labels (\eg natural evolution, artificial corruption), while the latter one comes from the selection of model parameters and model structures (\eg \cite{behrmann2019invertible} proposes that invertible ResNet can be more generative for input samples from different distribution datasets).
Bayesian machine learning \cite{mackay1992information,mackay1992bayesian,mackay1992practical} which works with probabilistic models and uncertainty, defines probability distributions over functions and are used to learn the more and less likely ways to generalize from observed data. Existing uncertainty measurements \cite{maddox2019simple, gal2016dropout,hernandez2015probabilistic} which enhance the effectiveness and efficiency of Bayesian machine learning indeed achieve some progress.



\subsection{Problem Definition}

\input{Table/dataset1}
\input{Table/dataset2}
\input{Table/dataset3}

Let $x\in\mathbb{R}^{n}$ denotes a space of $n$-dimensional features and $y \in \left \{ 1,\cdot \cdot \cdot ,C \right \}$ be its label for $C$-class classification. Suppose that a training dataset $D$ is generated from an unknown true distribution $p^{*}(x,y)$ and contains $M$ i.i.d. samples $D=\left \{ (x_{m}, y_{m}) \right \}_{m=1}^{M}$, also known as the \textit{data generation process} \cite{ovadia2019can}. In our programming language classification tasks, the true distribution is proposed to be a discrete distribution over the $C$ classes, and the observed label $y\in \left \{ 1,\cdot \cdot \cdot ,C \right \}$ is sampled from the conditional distribution $p^{*}(y|x)$. Let $f(\cdot)=p_{\theta }(y|x)$ denotes a neural network estimating the parameter $\theta$ through the training dataset.
In the testing phase, we evaluate the model predictions on a test dataset that is sampled from the same distribution as the training dataset and also on test datasets that are sampled from a shifted distribution $\hat{p}(x,y)\neq p^{*}(x,y)$. Note that in CV and NLP tasks, they typically consider \textit{a complete different OOD dataset} where the ground truth label in test dataset is not one of the $C$ classes, while in the programming language task we generally consider the \textit{shifted dataset} where the ground truth label belongs to one of the $C$ classes. This is because in the program dataset, only the variable names are different and user-defined while the special and key tokens are the same across different code snippets. In the next section, we introduce three types of distribution shift on the Java-based program dataset and evaluate the pre-trained model on the test dataset with different distribution shift.

%% file: Table/dataset1.tex
\begin{table*}[ht]
\caption{Dataset for different versions (timelines)}
\resizebox{1\textwidth}{!}{\begin{tabular}{c|ccccccc|c}

\toprule
\multirow{2}{*}{Dataset} & \multicolumn{7}{c|}{Version} & \multirow{2}{*}{Release Time} \\ 
& Elasticsearch & Gradle & Presto & Wildfly & Hadoop & Hibernate-orm & Spring-framework  \\
\midrule
Train & v.0.90.6  & REL\_1.9 & 0.53  & 8.0.0.Beta1 & REL\_2.2.0 &  4.3.0.CR1 & v.3.2.5.REL & Nov 2013\\

Test1 & v.2.3.2 & REL\_2.13 & 0.145 & 10.1.0.CR1 & YARN-2928 & 4.2.23.Final & v.3.2.17.REL & Apr 2016 \\

Test2 & v.6.6.2 & v.5.3.0 & 0.220 & 17.0.0.Alpha1 & OZONE-0.4.0 & 5.3.0 & v.5.2.0.M2 & Apr 2019 \\

Test3 & v.7.11.1 & v.6.8.3 & 0.248 & 23.0.0.Beta1 & REL\_3.2.2 & 5.4.29 & v.5.3.4 & Feb 2021 \\
\bottomrule
\end{tabular}}
\label{different_time}
\end{table*}

%% file: Table/dataset2.tex
\begin{table}[ht]
\caption{Dataset for different projects}
\resizebox{0.49\textwidth}{!}{\begin{tabular}{p{1cm}p{1cm}|cc|c}

\toprule
\multicolumn{2}{c|}{Dataset}   & Project & Version & Release Time \\

\midrule
\multirow{ 2}{*}{Dataset1} & Train & Spring-framework & v.5.3.4 & \multirow{ 2}{*}{Feb 2021} \\
& Test & Gradle & v.6.8.3 \\

\midrule
\multirow{ 2}{*}{Dataset2} & Train & Hibernate-orm & 5.4.29 & \multirow{ 2}{*}{Feb 2021}  \\
& Test & Hadoop & REL\_3.2.2 \\

\midrule
\multirow{ 2}{*}{Dataset3} & Train & Presto & 0.248 & \multirow{ 2}{*}{Feb 2021}  \\
& Test & Elasticsearch & v.7.11.1 \\

\bottomrule
\end{tabular}}
\label{different_project}
\end{table}


%% file: Table/dataset3.tex
\begin{table}[ht]
\caption{Dataset for different authors}
\resizebox{0.49\textwidth}{!}{\begin{tabular}{c|c|cc|c}

\toprule
Dataset & Author & Project & Version & Release Time \\

\midrule
Train & jasontedor & \multirow{ 4}{*}{Elasticsearch} & \multirow{4}{*}{5.4.29} & \multirow{4}{*}{Feb 2021} \\
Test1 & martijnvg & & \\
Test2 & s1monw & &\\
Test3 & kimchy & &\\

\bottomrule
\end{tabular}}
\label{different_author}
\end{table}

%% file: Distribution.tex
\section{Distribution shift in programming tasks}

\label{sec:shift_effect}



In this section, we propose three real-world program project dataset distribution shift scenarios and build three corresponding shifted datasets, then we implement two common programming language classification tasks and evaluate with the three shifted datasets. 
For each program task we follow standard training, validation and testing protocols except that we additionally evaluate results on increasingly shifted dataset. To show the correlation between different data distribution shift and model performance, we first illustrate our datasets configuration in Section~\ref{shift_dataset} and then introduce two downstream programming language tasks and report the model's prediction accuracy under different shifted datasets in Section~\ref{program_tasks} to illustrate the pattern of distribution shift.

\subsection{Research Questions}
We aim to answer the following two research questions in this section:

\noindent\textbf{\textit{RQ1.1}}:~Does distribution shift of program data affect the DL model performance? And how much impact each type of distribution shift has on the effectiveness of DL model?

\noindent\textbf{\textit{RQ1.2}}:~What factors may decide the effect of distribution shift on DL model performance?

\subsection{Datasets configuration}
\label{shift_dataset}

Suppose that a company designed a DL model to automatically check program misspell on a certain project $P$, as time goes this project has been committed for multiple times, \eg file addition and deletion, code modification, etc., and updated to a newer version which we denote as $P'$. It is critical to investigate the pre-trained model performance on the project $P'$ to ensure DL software's reliability on distribution shift across \textit{different timelines}. On the other hand, the company may also want to know if the pre-trained model can work well on other projects so that they can save unnecessary training effort, which we denotes as the distribution shift across \textit{different projects}. Furthermore, some of the new commits on project $P$ are done by other employees that do not originally participate in this project. Given the observation that different programmers may have different program writing habits, these commits implemented by new authors may also bring distribution shift and it is also necessary to evaluate the pre-trained model performance on the code snippets written by \textit{different authors}.

To simulate the three program data distribution shift scenarios, we pull 7 Java projects from GitHub, namely, \textit{elasticsearch} \cite{elasticsearch}, \textit{gradle} \cite{gradle}, \textit{presto} \cite{presto}, \textit{wildfly} \cite{wildfly}, \textit{hadoop} \cite{hadoop}, \textit{hibernate-orm} \cite{hibernate-orm}, \textit{spring-framework} \cite{spring-framework} and extract all Java files for the later programming tasks. For the first scenario, we choose four release time periods for all the 7 projects to represent the distribution shift across timelines. We combine all 7 projects released at each timeline to be the training, test1, test2 and test3 dataset, respectively. Intuitively, the degree of distribution shift is increasing as the time span enlarges. The detailed project versions and release timelines are shown in Table~\ref{different_time}. To present the second scenario, we choose three pairs of different projects that are all in the latest version. For each pair of dataset, one project is used for training and the other is for testing purpose. The detailed dataset configurations including project, version and release time are shown in Table~\ref{different_project}. For the third scenario, we investigate a specific project \textit{elasticsearch} by partitioning its Java files into four parts where files in each part are written by an author. Note that the project we choose is relatively large and contains hundreds of contributors, so we choose the four authors who have the most commits in history and the corresponding Java files as the dataset. The detailed information about authors is shown in Table~\ref{different_author}.

To visualize the distribution shift in programming data, we select one specific function (named \textit{extends IMutation}) and show its two versions written by author \textit{jasontedor} and \textit{kimchy} in Listing~\ref{lst:listing1} and Listing~\ref{lst:listing2}, respectively. We omit some parts of the first program for saving space. From this example we can conclude that different authors may write totally different programs even for realizing the same functionality, which incurs program distribution shift. 

\input{Figure/listing}

\subsection{Downstream Tasks}

\label{program_tasks}

In this section, we demonstrate the effect of data distribution shift on model performance under two programming language tasks that target different properties of source code: code summarization (CS) \cite{alon2019code2vec} and code completion (CC). They are both implemented using the datasets in Section~\ref{shift_dataset}. For clarification, there are other programming language tasks such as authorship identification, API search or code clone detection, but the datasets used in these tasks requires large labelling effort. Since our study focus on distribution shift and our shifted data are manually generated, we only consider CS and CC in this paper.

\noindent\textbf{Code Summarization.} The first task is source code summarization and more specifically, we consider predicting method names according to method bodies. We follow \cite{alon2019code2vec} to configure a path-attention network architecture for code prediction tasks and evaluate by the accuracy metric following work \cite{allamanis2016convolutional}.

\noindent\textbf{Code Completion.} The second task is to predict the missing code based on existing context. We follow \cite{rong2014word2vec} to configure a multilayer perceptron (MLP) architecture for code prediction tasks and apply accuracy as the evaluation metric.

\subsection{Experiment Settings}
\label{sec:model_setting}

We describe how we train the two task models on which we will evaluate their performance:

The path-attention model \cite{alon2019code2vec} (also refer to as \textit{Code2Vec}) for CS task contains two embedding layers (node\_embedding, path\_embedding), one dropout layer, one attention layer and one fully-connected (FC) layer. The MLP model \cite{rong2014word2vec} (also refer to as \textit{Word2Vec}) for CC task consists of one embedding layer and one FC layer. Their model parameters are shown in Table~\ref{model_params}.

\input{Table/parameters}

\subsection{Evaluation Results}

\input{Table/model_performance}

The model prediction accuracy (\%) is shown in Table~\ref{model_performance}. We also report the accuracy drop ratio for each test dataset to illustrate the impact of distribution shift on model performance.

\noindent
\\
\fbox{%
  \parbox{0.475\textwidth}{%
      \textbf{Finding 1}: \textit{All three types of code data distribution shift can  degrade the DL model performance in programming language tasks.}
  }%
}\\[0.1pt]

In both CS and CC tasks, the DL model performance is declined on each shifted test dataset compared to the validation accuracy under the three shift types. However, among the three types of program shift, distribution shift across timeline (version) has little effect on the DL performance, which presents the robustness of DL models under certain program data shift assumptions.

\noindent
\\
\fbox{%
  \parbox{0.475\textwidth}{%
      \textbf{Finding 2}: \textit{The degree of data distribution shift decides the impact of distribution shift on model performance.}
  }%
}\\[0.1pt]

As we see in Table~\ref{model_performance}, distribution shift across different timelines has pretty mild effect on both task performance, but the prediction accuracy on test1, test2, test3 dataset decrease in order as the corresponding release time span (degree of distribution shift) grows up, which means the severer shift we have on test dataset, the poorer DL model performance will be. Moreover, for distribution shift across different authors, the drop ratio of test1 dataset is much lower than that in test2 and test3 dataset, which indicates that author \textit{jasontedor} and author \textit{martijnvg} may have similar writing habits and the two data distributions are similar.

\noindent
\\
\fbox{%
  \parbox{0.475\textwidth}{%
      \textbf{Finding 3}: \textit{The type of data distribution shift also decides the impact of distribution shift.}
  }%
}\\[0.1pt]

From Table~\ref{model_performance} we can see the overall drop ratios of model prediction accuracy on the shifted test datasets under the three shift types are totally different. With distribution shift across different timelines having the slightest effect where the reduction of accuracy is less than 3\%, distribution shift across different projects having the severest impact where the maximum deduction of prediction accuracy reaches 57\% on Dataset3. This actually fits in the practical scenarios where most of the Java files in the same project are unchanged through timelines, but different projects or projects written by different authors may contain totally different composition and distribution of program snippets.

\noindent
\\
\fbox{%
  \parbox{0.475\textwidth}{%
      \textbf{Finding 4}: \textit{The impact of data distribution shift on DL model performance also relies on the model architecture. Simple DNN architecture is more robust to data distribution shift.}
  }%
}\\[0.1pt]

Comparing the overall drop ratios under the three shift types on both CS and CC tasks, we find that the impact of distribution shift is more obvious on CS than CC for each dataset, \eg the drop ratios under author shift for test1, test2 and test3 dataset are 0.96\%, 45.40\%, 48.95\% for CS and 0.29\%, 6.32\%, 7.11\% for CC. We assume that the two DL models under both CS and CC tasks are well-trained, then the different impact of distribution shift on the classification task mainly raises from the different model architectures which have different sensitivity to distribution shift. Note that the DL model architecture in CC is just one FC-layer MLP which is much shallower and simpler than the path-attention network architecture in CS. Given that larger number of parameters in a more complex and deeper neural network are more likely to cause overfitting in the training data distributing, we can conclude that a simple DNN architecture should be more robust to distribution shift.

%% file: Figure/listing.tex
\definecolor{codegreen}{rgb}{0,0.6,0}
\definecolor{codegray}{rgb}{0.5,0.5,0.5}
\definecolor{codepurple}{rgb}{0.58,0,0.82}
\definecolor{backcolour}{rgb}{0.99,0.99,0.99}

\lstdefinestyle{mystyle}{
  backgroundcolor=\color{backcolour},   commentstyle=\color{codegreen},
  keywordstyle=\color{magenta},
  numberstyle=\tiny\color{codegray},
  stringstyle=\color{codepurple},
  basicstyle=\ttfamily\footnotesize,
  breakatwhitespace=false,         
  breaklines=true,                 
  captionpos=b,                    
  keepspaces=true,                 
  numbers=left,                    
  numbersep=5pt,                  
  showspaces=false,                
  showstringspaces=false,
  showtabs=false,                  
  tabsize=2
}

\lstset{style=mystyle}

\begin{lstlisting}[language=JAVA,
label={lst:listing1},
caption=Java code written by author \textit{jasontedor}]
private Collection<? extends IMutation> getMutations(BatchQueryOptions options, boolean local, long now)
    throws RequestExecutionException, RequestValidationException
    {
        Set<String> tablesWithZeroGcGs = null;
        UpdatesCollector collector = new UpdatesCollector(updatedColumns, updatedRows());
        for (int i = 0; i < statements.size(); i++)
        {
            ... // omitted part
        }
        if (tablesWithZeroGcGs != null)
        {
            ... // omitted part
        }
      collector.validateIndexedColumns();
        return collector.toMutations();
    }
\end{lstlisting}

\begin{lstlisting}[language=JAVA,
label={lst:listing2}, caption=Java code written by author \textit{kimchy}]
private Collection<? extends IMutation> getMutations(QueryOptions options, boolean local, long now)
    {
        UpdatesCollector collector = new UpdatesCollector(Collections.singletonMap(cfm.cfId, updatedColumns), 1);
        addUpdates(collector, options, local, now);
        collector.validateIndexedColumns();
        return collector.toMutations();
    }

\end{lstlisting}

%% file: Table/parameters.tex
\begin{table}[H]
\caption{Parameters for Code2Vec and Word2Vec models}
\resizebox{0.49\textwidth}{!}{\begin{tabular}{ccc}

\toprule
Parameter & Code2Vec Model & Word2Vec Model \\

\midrule
Learning Rate & 0.001 & 0.001 \\
Embedding Dimension & 100 & 100 \\
Dropout & 0.5 & - \\
Optimizer & Adam & Adam \\
Batch Size & 512 & 512 \\
Epochs & 300 & 300\\

\bottomrule
\end{tabular}}
\label{model_params}

\end{table}

%% file: Table/model_performance.tex
\begin{table*}[ht]
\caption{Programming task performance (accuracy) on 3 types of data distribution shift}
\resizebox{1\textwidth}{!}{\begin{tabular}{c|cccc|cccc}

\toprule
Shift Type & \multicolumn{4}{c|}{CS} & \multicolumn{4}{c}{CC} \\

\midrule
\multirow{2}{*}{Timeline} & Validation & Test1 & Test2  & Test3 & Validation & Test1 & Test2  & Test3 \\
& 29.96 & 29.86(-0.33\%) & 29.7(-0.87\%) & 29.14(-2.74\%) & 45.45 & 45.4(-0.11\%) & 44.9(-1.21\%) & 44.29(-2.55\%) \\

\midrule
\multirow{3}{*}{Project} & & Dataset1 & Dataset2 & Dataset3 & & Dataset1 & Dataset2 & Dataset3 \\
& Validation & 28.47 & 55.02 & 47.39 & Validation & 48.35 & 50.83 & 50.44 \\
& Test & 27.25(-4.29\%) & 28.38(-48.42\%) & 20.11(-57.56\%)  & Test & 40.13(-17.00\%) & 39.15(-22.98\%) & 40.10(-20.50\%) \\

\midrule
\multirow{2}{*}{Author} & Validation & Test1 & Test2  & Test3 & Validation & Test1 & Test2  & Test3 \\
& 45.66 & 45.22(-0.96\%) & 24.93(-45.40\%) & 23.31(-48.95\%) & 47.81 & 47.67(-0.29\%) & 44.79(-6.32\%) & 44.41(-7.11\%) \\

\bottomrule
\end{tabular}}
\label{model_performance}
\end{table*}

%% file: Metric.tex
\section{Uncertainty Estimation}
\label{sec:uncertainty_eval}

\input{Table/Uncertainty_correctness}

We implement our evaluation of the existing popular uncertainty measurements on our previously defined datasets and well-trained models in Section~\ref{sec:shift_effect}. For each program task we evaluate the effectiveness of the following popular uncertainty methods through multiple metrics in terms of both error/success prediction and in-/out-of-distribution detection.

\subsection{Research Questions}
We aim to answer the following four research questions in this section:

\noindent\textbf{\textit{RQ2.1}}:~How effective are the existing uncertainty methods in error/success prediction on shifted program data? Which method(s) perform(s) relatively well (or bad) and why?

\noindent\textbf{\textit{RQ2.2}}:~How effective are the existing uncertainty methods in distinguishing in/out-of-distribution program inputs? Which method(s) perform(s) relatively well (or bad) and why?

\noindent\textbf{\textit{RQ2.3}}:~Are the existing uncertainty methods sensitive to program distribution shift?

\noindent\textbf{\textit{RQ2.4}}:~Are the performance of existing uncertainty methods consistent on CS and CC? If not consistent, what are the possible reasons?

\subsection{Uncertainty Methods}

In recent years, lots of researches about uncertainty measurement for DL models have been proposed. We select a subset of uncertainty metrics from the existing literature for their prevalence, scalability, and practical applicability. The selected work includes:


\begin{itemize}
    \item\textit{\textbf{Vanilla}} \cite{hendrycks2016baseline}: \textit{Vanilla} proposes that the maximum softmax probability could be used as the model confidence or uncertainty to distinguish the in- and OOD inputs.

    \item\textit{\textbf{Temp Scale}} \cite{guo2017calibration}: \textit{Temp scale} utilizes the post-hoc calibration named temperature scaling on a validation set to calibrate the gap between model predictive confidence and accuracy, the calibrated DL model's predictive probability could represent better ground-truth correctness likelihood and thus be used for uncertainty measurement.
    
    \item\textit{\textbf{MC-Dropout}} \cite{gal2016dropout}: \textit{Monte-Carlo Dropout} with rate $p$ as approximate Bayesian inference in deep Gaussian process. It shows the dropout process can distinguish between in- and OOD samples.

    \item\textit{\textbf{mMutate}} \cite{wang2019adversarial}: \textit{mMutant} integrates statistical hypothesis testing and model mutation testing to check whether an input sample is likely to be in- or OOD at runtime by measuring its sensitivity to the model mutation. Based on different mutation operators, mMutant is configured into four sub-techniques, named as, \textit{mMutant-GF}, \textit{mMutant-NAI}, \textit{mMutant-WS} and \textit{mMutant-NS}. They evaluate the uncertainty in terms of \textit{Label Changing Rate (LCR)}.

    
    \item\textit{\textbf{Dissector}} \cite{wang2020dissector}: \textit{Dissector} proposes a model-specific uncertainty evaluation approach based on assessing the model’s cross-layer confidence about a given input. Based on different weight growth types, Dissector was configured into three sub-techniques, named as, \textit{Dissector-linear}, \textit{Dissector-log} and \textit{Dissector-exp}.

\end{itemize}

\subsection{Uncertainty Experimental Setup}
\label{sec:uncertainty_setup}

\subsubsection{Datasets and Models}~We continue using the datasets and well-trained DL models in Section~\ref{sec:shift_effect}.

\subsubsection{Uncertainty Methods Setup}~For \textit{Vanilla}, no additional configuration is needed since it leverages the softmax probability of the DL models to measure predictive uncertainty. 
For \textit{Temp Scale}, we use the \textit{BFGS} optimizer \cite{battiti1990bfgs} to train the calibration temperature on additional validation dataset which has the same distribution as the training set with a learning rate of 0.01.
For \textit{MC-Dropout}, we follow \cite{gal2016dropout} and set dropout probability as 0.5. For \textit{mMutate}, we follow \cite{wang2019adversarial} and use four mutation operators, namely, Gaussian Fuzzing (GF), Weight Shuffling (WS), Neuron Switch (NS) and Neuron Activation Inverse (NAI), and report the best score of the four in our experimental results. We also configure the mutation degree as 0.05 in our study.
For \textit{Dissector}, it needs to select interior NN layers for sub-model generation. For the path-attention model in CS task, we adapt the hidden feature from both embedding layer and attention layer. For the MLP model in CC task, we select the embedding layer of each token as the feature to train the sub-model since there is only one FC layer. Similarly, we report the best evaluation scores of the three sub-techniques of \textit{Dissector} in Section~\ref{sec:uncertainty_res}.

\subsection{Evaluation Metrics}

We evaluate the above methods on both CS and CC tasks with our created datasets. We use arrows to indicate which direction is better for each of the metric. The evaluation metrics contains:

\begin{itemize}

    \item\noindent\textit{\textbf{AUC}}$\uparrow$\cite{wang2019adversarial}: the Area Under the Receiver Operating Characteristic curve (AUC) is a threshold-independent performance evaluation metric \cite{davis2006relationship}. The ROC curve is a figure showing the relation between true positive rate and false positive rate. AUC represents the probability that a positive example has a larger predictive score than a negative one \cite{fawcett2006introduction}. A random classifier corresponds to an AUC of 50\% and a "perfect" classifier corresponds to 100\%. 
    
    \item\noindent\textit{\textbf{AUPR}}$\uparrow$\cite{manning1999foundations}: the Area Under the Precision-Recall curve (AUPR) better handles the situation when the positive class and negative class have greatly differing base rates compared to AUC. The PR curve plots the relationship between precision and recall. The baseline classifier has an AUPR roughly equal to the precision \cite{saito2015precision} and a "perfect" classifier corresponds to an AUPR of 100\%.
    
    \item\noindent\textit{\textbf{Brier Score}}$\downarrow$\cite{brier1950verification}: proper scoring rule representing the accuracy of predicted probabilities. It measures the mean squared error of the predicted probability assigned to the possible outcomes for each sample and the actual outcome. Therefore, the lower the Brier score is for a set of predictions, the better the predictions are calibrated. 

\end{itemize}

\subsection{Evaluation Results}
\label{sec:uncertainty_res}

\input{Figure/cs1}

\input{Figure/cc1}

\subsubsection{Error/Success Prediction}

Table~\ref{Correct_eval} shows the error/success prediction results of the 5 uncertainty metrics all the validation sets. We mark the best metric score of the 5 uncertainty methods in \textbf{bold} format. Fig.~\ref{cs1} and Fig.~\ref{cc1} respectively show the uncertainty evaluation on CS and CC tasks under not only the validation set but also shifted datasets.

\textbf{Ground truths.} The ground truths of uncertainty prediction are inputs that the model can handle (within inputs) and those the model cannot handle (beyond inputs). Based on our previous discussion that beyond-inputs are likely to cause a DL model's prediction to be misleading or wrong. Therefore, we consider within inputs (positive) as those correctly-predicted samples and beyond inputs (negative) as those incorrectly-predicted samples.

\noindent
\\
\fbox{%
  \parbox{0.475\textwidth}{%
      \textbf{Finding 5}: \textit{The effectiveness of softmax-based uncertainty estimates highly rely on the DL model performance.}
  }%
}\\[0.1pt]

In Table~\ref{model_performance}, the model's prediction accuracy on validation set are relatively low in CS task under timeline shift and under project shift dataset1. In both cases, the model's prediction accuracy is less than 30\%. Correspondingly in Table~\ref{Correct_eval}, the overall performance of \textit{Vanilla} and \textit{Temp Scale} in these two situations are also relatively poor compared to \textit{MC-Dropout} and \textit{Dissector}. In comparison, the model's prediction accuracy in CC task are more decent with most values larger than 40\%, the corresponding \textit{Vanilla} and \textit{Temp Scale} uncertainty evaluation scores are much better than other uncertainty methods. One reason is that \textit{Vanilla} and \textit{Temp scale} both focus on leveraging the DNN's softmax layer to represent the confidence score, given that softmax layer is the output layer that is directly correlated with the model's prediction, their calibrated confidence score has more reliance on the original model's prediction accuracy than those uncertainty methods that focus more on the DNN's interior layer activation values.

\noindent
\\
\fbox{%
  \parbox{0.475\textwidth}{%
      \textbf{Finding 6}: \textit{Layer-level uncertainty estimates perform well in complex or deep neural network but struggle with shallow DL model.}
  }%
}\\[0.1pt]

In Table~\ref{Correct_eval} we find that \textit{MC-Dropout} and \textit{Dissector} perform well in CS task under nearly all different datasets but perform poorly in CC task under all cases. This is because the two methods are based on calculating the consistency between network units in each layer, and these layer-level predictive uncertainty methods require deeper neural network architecture to ensure fidelity. Since the MLP model used in CC task contains only one fully-connected (FC) layer, all uncertainty estimates of \textit{MC-Dropout} and \textit{Dissector} under CC task get compromised. In comparison, the path-attention model used in CS task contains 4 interior layers, which provides more versatile cross-layer confidence about a given input and thus makes the calibrated uncertainty of the layer-level uncertainty methods more reliable.

\noindent
\\
\fbox{%
  \parbox{0.475\textwidth}{%
      \textbf{Finding 7}: \textit{Adversarial-based uncertainty estimates can hardly detect beyond inputs from an in-distribution dataset.}
  }%
}\\[0.1pt]

Table~\ref{Correct_eval} demonstrates that \textit{mMutate}'s overall evaluation scores are relatively low, this results from that \textit{mMutate} is proposed for detecting adversarial samples that are closer to the model decision boundary and sensitive to the mutation operations. While in the error/success prediction case, all inputs are from the validation set and few of them are adversarial samples or out-of-distribution samples. Distinguishing adversarial inputs may not be enough for well-calibrating model's prediction confidence.

\noindent
\\
\fbox{%
  \parbox{0.475\textwidth}{%
      \textbf{Finding 8}: \textit{The overall evaluation scores on different metrics reflect the effectiveness of an uncertainty method rather than only a single metric.}
  }%
}\\[0.1pt]

In Table~\ref{Correct_eval} we mark the best evaluation score of the 5 uncertainty methods as \textbf{bold}. Experiments illustrate that the best evaluation scores of the three metrics not always belong to the same uncertainty method. For example, in CS task under timeline distribution shift, \textit{MC-Dropout} has the best AUC of 75.05, \textit{mMutate} has the best AUPR of 69.89 while \textit{Vanilla} has the best Brier of 27.98. This is due to that different metrics evaluate different aspect of prediction accuracy. Particularly, Brier score measures the marginal uncertainty over labels and are insensitive to predicted probabilities associated with in/frequent events, thus \textit{Temp Scale} which calibrates the confidence scores into a small interval with high frequency can achieve decent Brier score; AUPR compromises with the positive and negative label rates, consequently, the imbalanced ground truths in CS task under different timelines and under different projects dataset1 due to the low model prediction causes the AUPR to be much lower than AUC. Therefore, one should evaluate the effectiveness of an uncertainty based on different metric scores.

\noindent
\\
\fbox{%
  \parbox{0.475\textwidth}{%
      \textbf{Finding 9}: \textit{Layer-level uncertainty methods are more robust to dataset shift on CS task, while softmax-based uncertainty methods are more robust to dataset shift on CC task.}
  }%
}\\[0.1pt]

Generally, a model uncertainty that is well-calibrated on the training and validation distributions would ideally remain so on shifted dataset \cite{ovadia2019can}. But as shown in Fig.~\ref{cs_author} and Fig.~\ref{cc_author}, all the five uncertainty methods exhibit varying degrees of AUC and Brier decline as the model's prediction accuracy decreases on the increasing shifted test dataset, which indicates that existing uncertainty metrics still need further improvement for programming language applications. We also explore that layer-level uncertainty such as \textit{MC-Dropout} and \textit{Dissector} are more robust to dataset shift on code summary task as evidenced by a lower overall confidence Fig.~\ref{cs_project1}, Fig.~\ref{cs_project2}, Fig.~\ref{cs_project3} (the number of samples that are larger than the confidence threshold are smaller than other methods) and higher overall AUC and lower Brier as shown in Fig.~\ref{cs_time1}, Fig.~\ref{cs_time2}, Fig.~\ref{cs_author}, while softmax-based uncertainty such as \textit{Vanilla} and \textit{Temp Scale} are more robust to dataset shift on code completion task as evidenced by a lower overall confidence Fig.~\ref{cc_project1}, Fig.~\ref{cc_project2}, Fig.~\ref{cc_project3} and higher overall AUC and lower Brier as shown in Fig.~\ref{cc_time1}, Fig.~\ref{cc_time2}, Fig.~\ref{cc_author}. This finding is actually consistent with our previous findings that softmax-based calibrated uncertainty works well on pre-trained models with high prediction accuracy, while layer-level uncertainty works well on deep and complex DL models (path-attention model in CS task).

\subsubsection{In-/out-of-distribution detection}

\input{Table/Uncertainty_ood_time}
\input{Table/Uncertainty_ood_project}
\input{Table/Uncertainty_ood_author}

Table~\ref{ood_eval_time}, Table~\ref{ood_eval_project}, Table~\ref{ood_eval_author} respectively shows the in-/OOD detection results of the 5 uncertainty metrics on CS and CC task under each type of distribution shift. For simplification, we denote "validation" as "val". We also mark the best metric score of the 5 uncertainty methods in \textbf{bold} format. Note that the distribution shift under programming data is relatively mild compared to those complete OOD datasets used in CV tasks, thus it is tougher for existing uncertainty methods to distinguish between in-/OOD datasets in our experiments which means lower evaluation scores in this part are still reasonable. 

\textbf{Ground truths.} The ground truths of in-/OOD detection are inputs that are within the training data distribution (in-distribution inputs) and those follow a different distribution from the training data (OOD inputs). Based on our setting, all test datasets contain more or less distribution shift should be considered as OOD data. Therefore, we consider in-distribution inputs (positive) as those samples in the validation set and OOD inputs (negative) as those samples in the shifted test sets. 

\noindent
\\
\fbox{%
  \parbox{0.475\textwidth}{%
      \textbf{Finding 10}: \textit{Existing predictive uncertainty methods are sensitive to data distribution shift, especially when the level of distribution shift is high.}
  }%
}\\[0.1pt]

From Table~\ref{model_performance} we know the degree of distribution shift across authors is increasing in order of test1, test2 and test3 as the prediction accuracy decreases, correspondingly in Table~\ref{ood_eval_author} the overall evaluation performance of the 5 uncertainty methods also becomes better in order of val/test1, val/test2 and val/test3. The pattern is same for distribution shift across projects as the drop ratio increases in the order of Dataset1, Dataset2 and Dataset3, the 5 uncertainty methods perform better in distinguish between the corresponding in/OOD dataset pair, \eg val1/test1, val2/test2 and val3/test3. This pattern demonstrates that existing predictive uncertainty are well-calibrated and sensitive to distribution shift. As the level of distribution shift increases, these uncertainty methods can more precisely distinguish between the in-distribution dataset and shifted OOD dataset.

\noindent
\\
\fbox{%
  \parbox{0.475\textwidth}{%
      \textbf{Finding 11}: \textit{Adversarial-based uncertainty estimates perform more effectively in distinguish in-/OOD inputs than predicting error/success inputs.}
  }%
}\\[0.1pt]

In comparison of the error/success prediction and in-/OOD detection results, we find that the overall performance of \textit{mMutate} over the other 4 uncertainty methods on OOD detection is relatively better than on error/success prediction, which further indicates that inputs from a shifted distribution are more sensitive to model mutation operations and this approach becomes more effective when detecting adversarial inputs or inputs from a different distribution.

\noindent
\\
\fbox{%
  \parbox{0.475\textwidth}{%
      \textbf{Finding 12}: \textit{Dissector is not sensitive to distribution shift and falls short in detecting program OOD inputs.}
  }%
}\\[0.1pt]

As shown in Table~\ref{ood_eval_time}, Table~\ref{ood_eval_project}, Table~\ref{ood_eval_author}, \textit{Dissector} struggles with distinguishing in-/OOD inputs in all three distribution shift cases. Specifically, most of its AUC and AUPR scores are around or below 50, which means the calibrated confidence scores on both validation set and shifted set are similar. This result also indicates that inputs from a different distribution shift are not necessary to cause the neural network's interior layer-level intermediate prediction inconsistency. And DL model's cross-layer confidence about a given inputs might be insensitive to data distribution shift.


%% file: Table/Uncertainty_correctness.tex
\begin{table*}[ht]
\caption{Uncertainty approach evaluation (error/success prediction)}
\resizebox{1\textwidth}{!}{\begin{tabular}{c|c|ccc|ccc|ccc|ccc|ccc}

\toprule
\multirow{3}{*}{Task} & \multirow{3}{*}{Approach} & \multicolumn{3}{c|}{Different Timelines} & \multicolumn{9}{c|}{Different Projects} & \multicolumn{3}{c}{Different Authors} \\
& & \multicolumn{3}{c}{} & \multicolumn{3}{|c}{Dataset1} & \multicolumn{3}{|c|}{Dataset2} & \multicolumn{3}{c|}{Dataset3} & \multicolumn{3}{c}{} \\
& & AUC & AUPR & Brier & AUC & AUPR & Brier & AUC & AUPR & Brier & AUC & AUPR & Brier & AUC & AUPR & Brier \\

\midrule
\multirow{5}{*}{CS} & Vanilla & 56.32 & 49.35 & \textbf{27.98} & 82.08 & 67.57 & 42.38 & 78.85 & 82.67 & 38.09 & 61.18 & 71.94 & 50.46 &  \textbf{87.11} & \textbf{85.61} & 24.00 \\
& Temp Scale & 56.42 & 49.42 & 28.02 & 78.87 & 62.79 & 28.43 & 51.70 & 77.83 & 54.88 & 50.06 & 73.70 & 47.30 & 86.94 & 85.30 & \textbf{15.83} \\
& mMutate & 67.28 & \textbf{69.89} & 46.02 & 83.63 & 72.99 & 28.61 & 78.55 & 85.37 & 26.62 & 81.69 & 81.67 & 22.77 & 74.33 & 78.55 & 37.37 \\
& MC-Dropout & \textbf{75.04} & 66.05 & 32.90 & \textbf{89.69} & \textbf{77.57} & \textbf{12.02} & 89.64 & 89.37 & \textbf{12.90} & 86.70 & 86.36 & \textbf{15.02} & 84.67 & 84.52 & 23.58 \\
& Dissector & 73.10 & 61.97 & 32.01 & 87.11 & 76.39 & 18.91 & \textbf{93.57} & \textbf{93.89} & 15.13 & \textbf{88.48} & \textbf{87.71} & 18.77 & 83.30 & 83.84 & 26.39 \\

\midrule
\multirow{5}{*}{CC} & Vanilla & \textbf{84.12} & 83.47 & 16.62 & \textbf{83.57} & \textbf{83.89} & 17.06 & \textbf{85.01} & \textbf{86.95} & 16.52 & \textbf{86.01} & \textbf{88.12} & 15.86 &  \textbf{85.15} & \textbf{85.44} & \textbf{16.02} \\
& Temp Scale & 84.00 & \textbf{83.59} & \textbf{16.23} & 83.30 & 83.78 & \textbf{16.92} & 84.72 & 86.87 & \textbf{16.11} & 85.75 & 87.89 & \textbf{15.54} & 84.84 & 85.43 & 16.54 \\
& mMutate & 67.44 & 72.86 & 43.81 & 63.04 & 74.80 & 44.48 & 66.64 & 78.16 & 38.90 & 60.80 & 76.11 & 44.97 & 64.66 & 73.50 & 44.36 \\
& MC-Dropout & 50.23 & 72.83 & 54.34 & 50.03 & 74.06 & 51.89 & 50.21 & 76.04 & 47.92 & 50.10 & 75.10 & 49.80 &  50.01 & 73.66 & 52.68 \\
& Dissector & 61.44 & 61.49 & 27.82 & 61.35 & 63.89 & 27.60 & 60.54 & 66.99 & 28.64 & 63.60 & 69.63 & 25.98 &  63.61 & 65.60 & 27.50 \\

\bottomrule
\end{tabular}}
\label{Correct_eval}
\end{table*}

%% file: Figure/cs1.tex
\begin{figure*}[htbp]
\centering

\subfloat[Timeline Shift (time view)]{\label{cs_time1}  \includegraphics[width=0.32\textwidth]{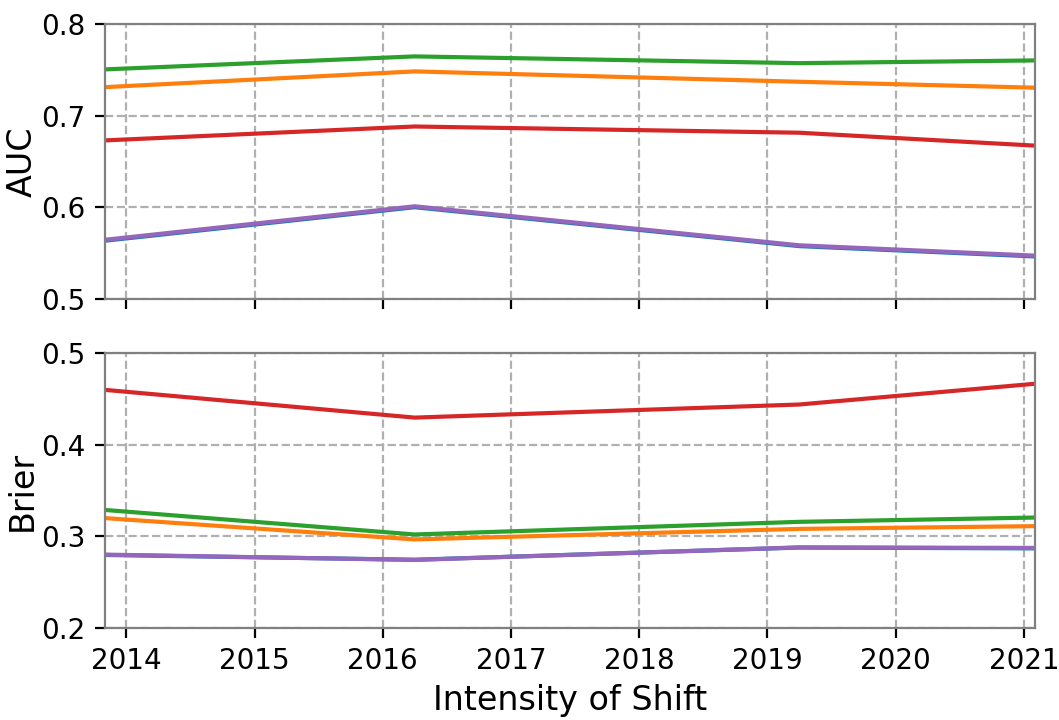}}
\subfloat[Timeline Shift (accuracy view)]{\label{cs_time2}  \includegraphics[width=0.32\textwidth]{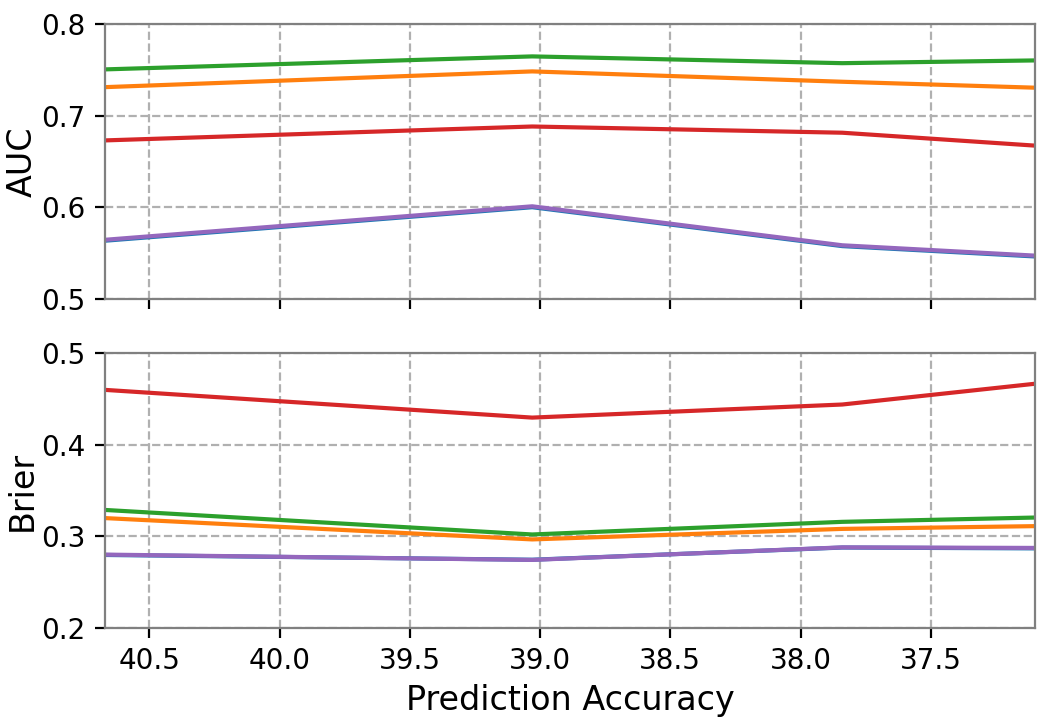}}
\subfloat[Author Shift (accuracy view)]{\label{cs_author}  \includegraphics[width=0.32\textwidth]{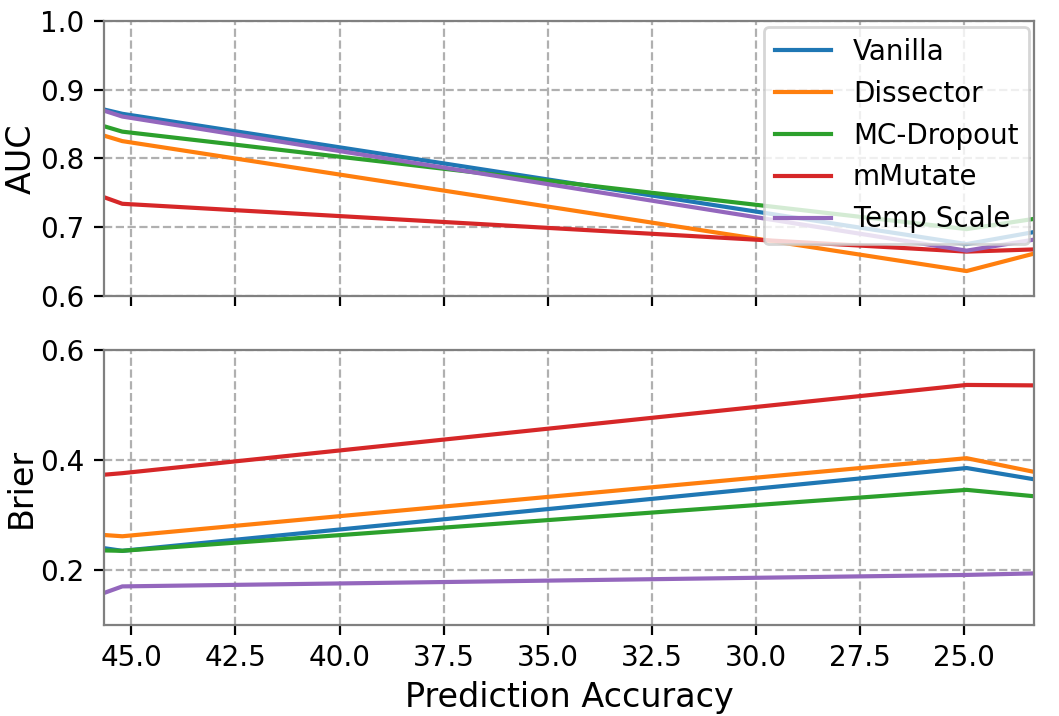}}\\
\subfloat[Project Shift1 ]{\label{cs_project1}  \includegraphics[width=0.32\textwidth]{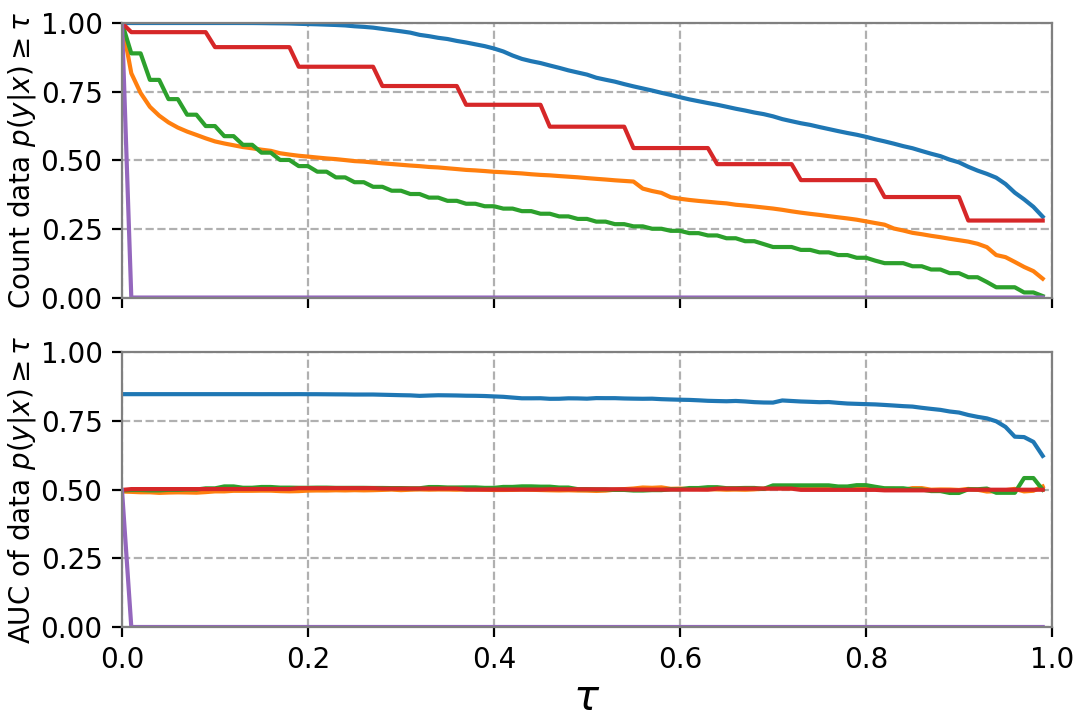}}
\subfloat[Project Shift2 ]{\label{cs_project2}  \includegraphics[width=0.32\textwidth]{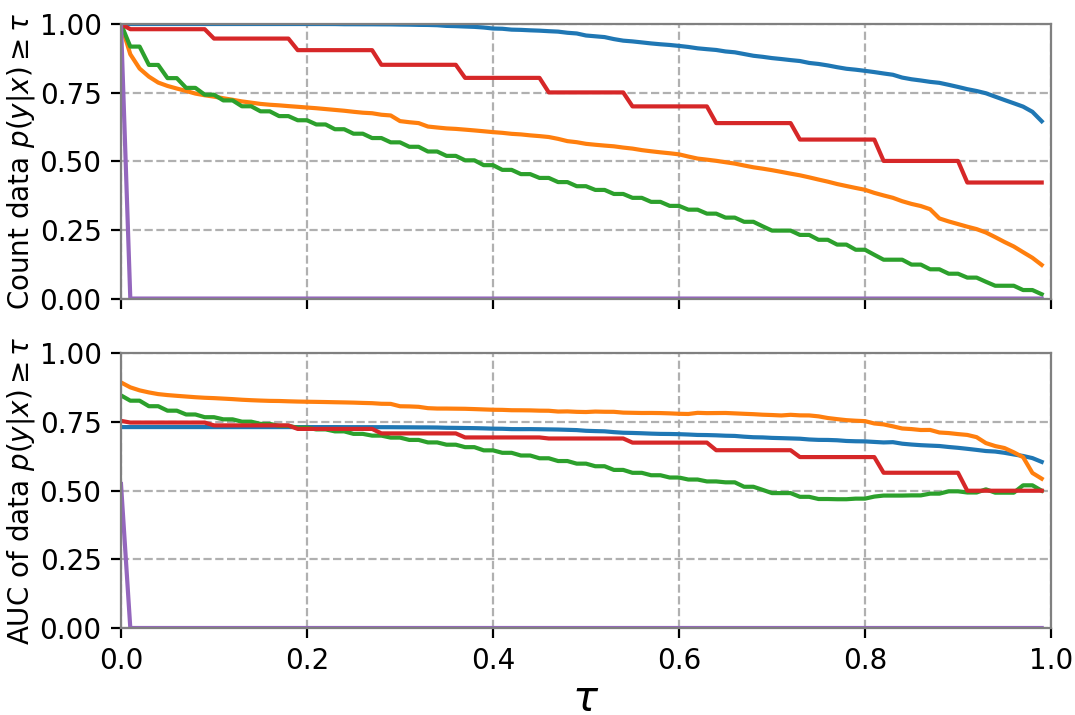}}
\subfloat[Project Shift3 ]{\label{cs_project3}  \includegraphics[width=0.32\textwidth]{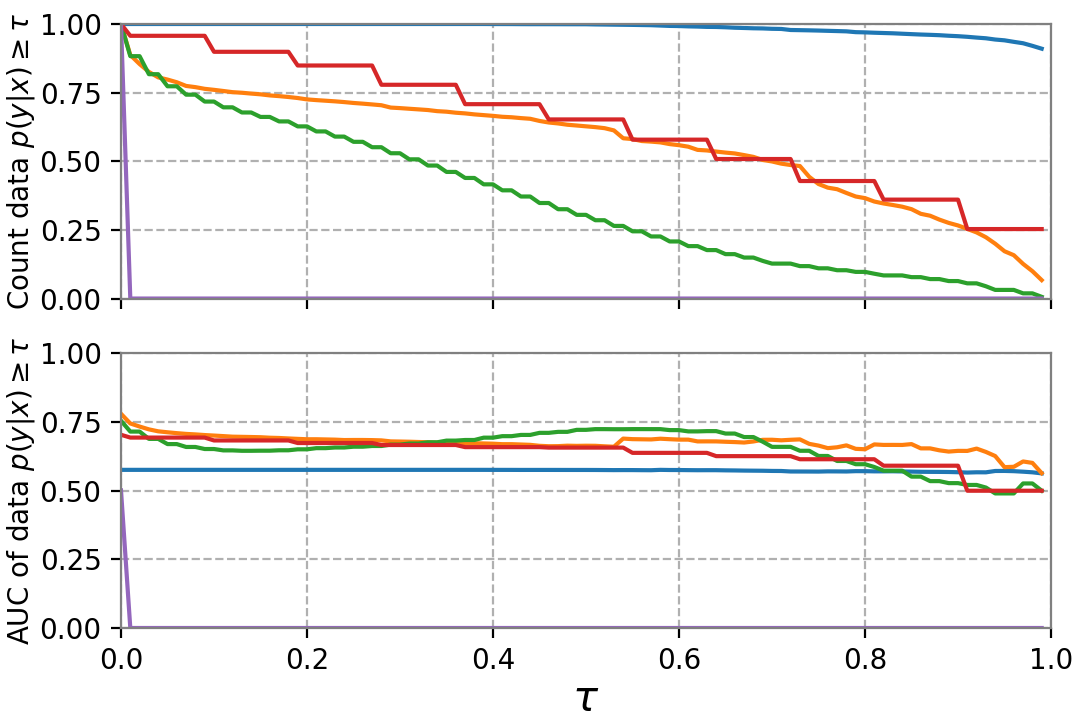}}

\caption{Results on Code Summary Task: \ref{cs_time1} and \ref{cs_time2} show AUC and Brier scores as the data is increasingly shifted (across timeline). \ref{cs_author} also presents the two scores as the data is increasingly shifted (across author). To investigate the effectiveness of uncertainty methods, we also explore the filtered Number of samples (Count) and AUC versus confidence score on the OOD data in \ref{cs_project1}, \ref{cs_project2} and \ref{cs_project3}. In the three project shift datasets, the number of samples decreases as the confidence score grows. \textit{Temp Scale} has lower AUC on three shifted projects since all of its confidence scores are in a very small range around 0. \textit{MC-Dropout} and \textit{Dissector} have the overall decent AUC and Brier. Their effectiveness is stable as the intensity of shift increases.}
\label{cs1}
\end{figure*}

%% file: Figure/cc1.tex
\begin{figure*}[htbp]
\centering

\subfloat[Timeline Shift (time view)]{\label{cc_time1}  \includegraphics[width=0.32\textwidth]{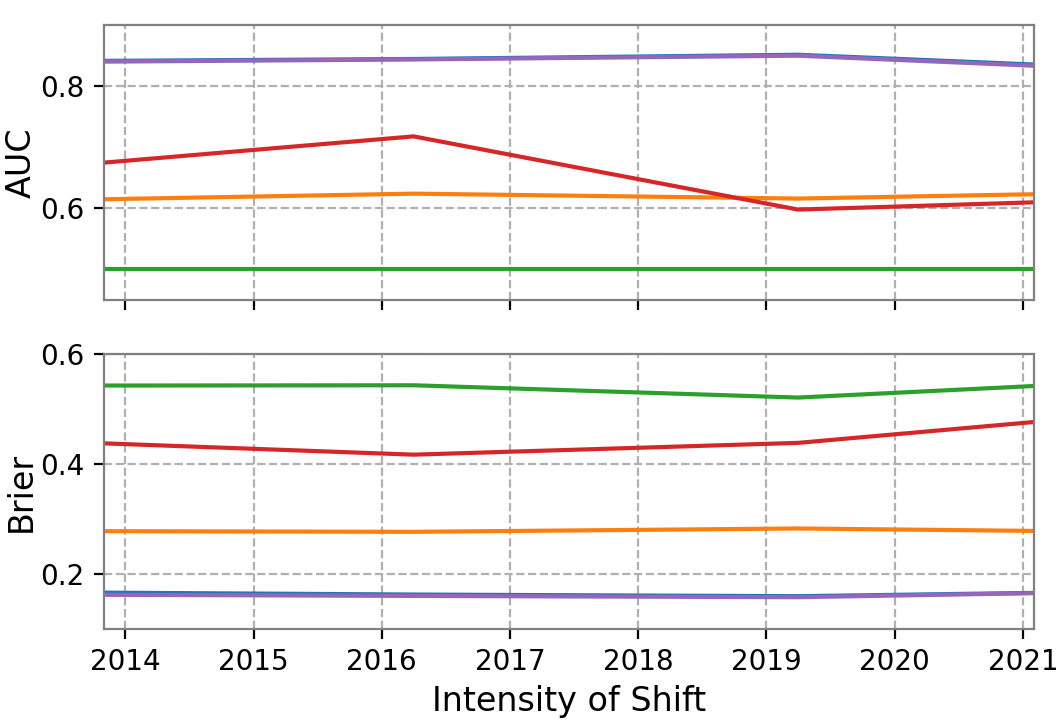}}
\subfloat[Timeline Shift (accuracy view)]{\label{cc_time2}  \includegraphics[width=0.32\textwidth]{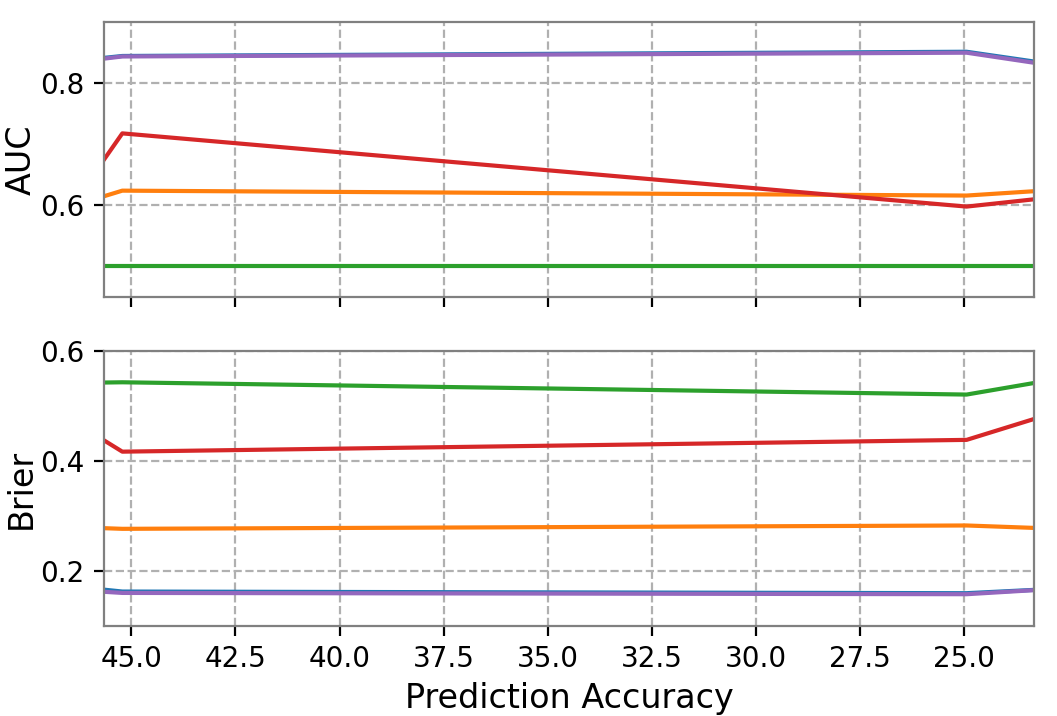}}
\subfloat[Author Shift (accuracy view)]{\label{cc_author}  \includegraphics[width=0.32\textwidth]{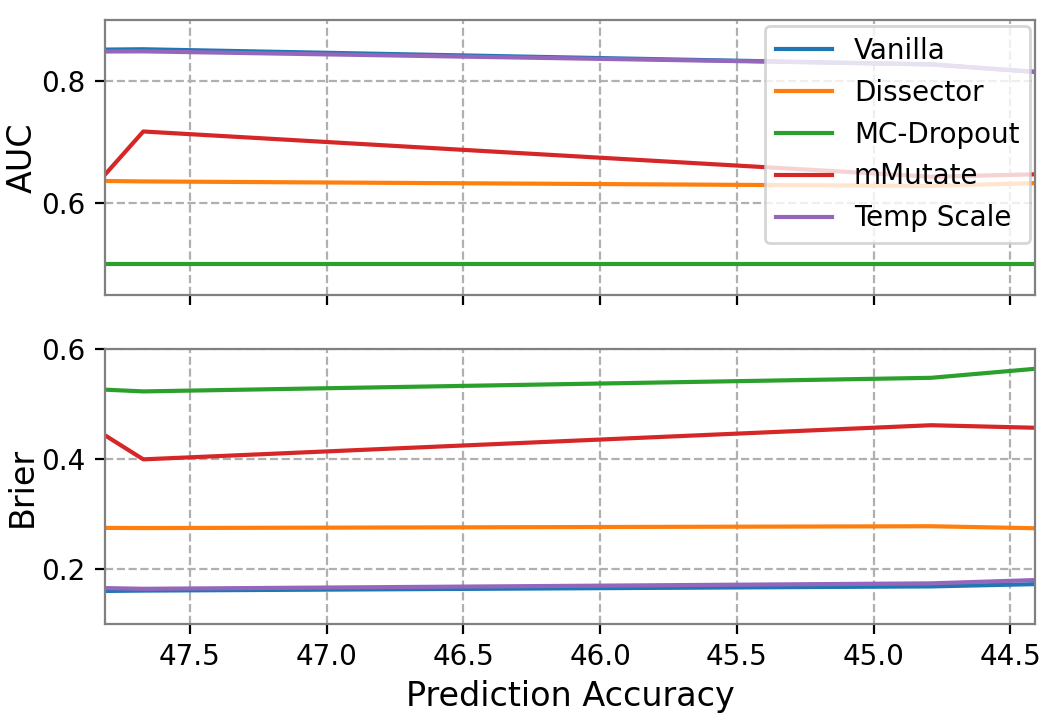}}\\
\subfloat[Project Shift1 ]{\label{cc_project1}  \includegraphics[width=0.32\textwidth]{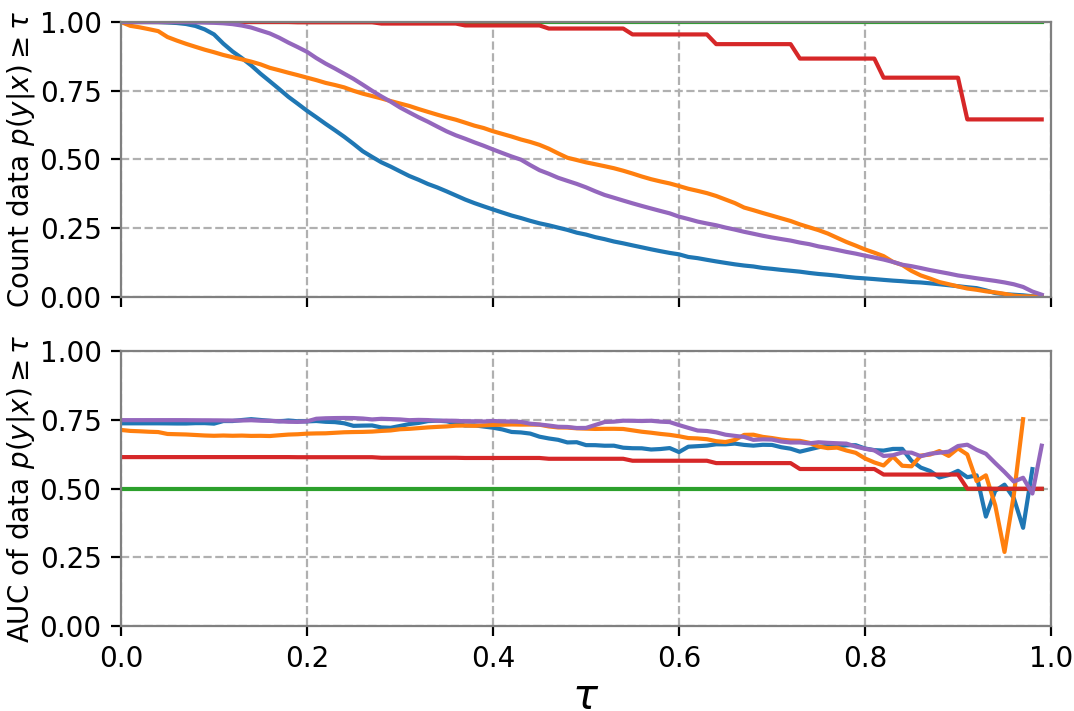}}
\subfloat[Project Shift2 ]{\label{cc_project2}  \includegraphics[width=0.32\textwidth]{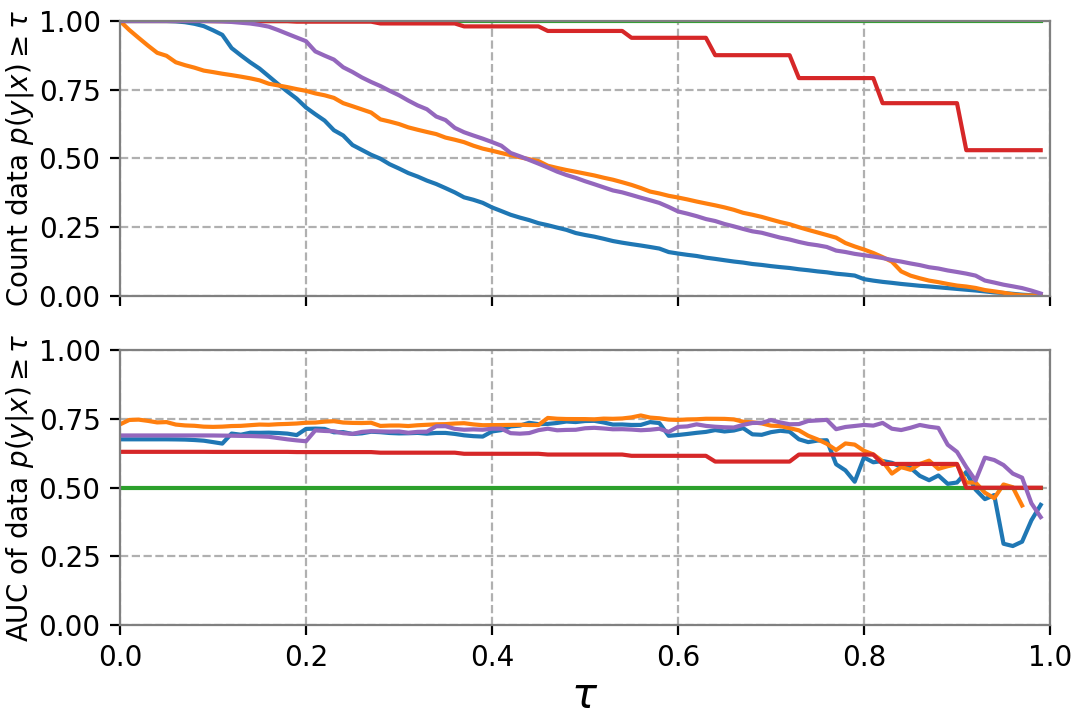}}
\subfloat[Project Shift3 ]{\label{cc_project3}  \includegraphics[width=0.32\textwidth]{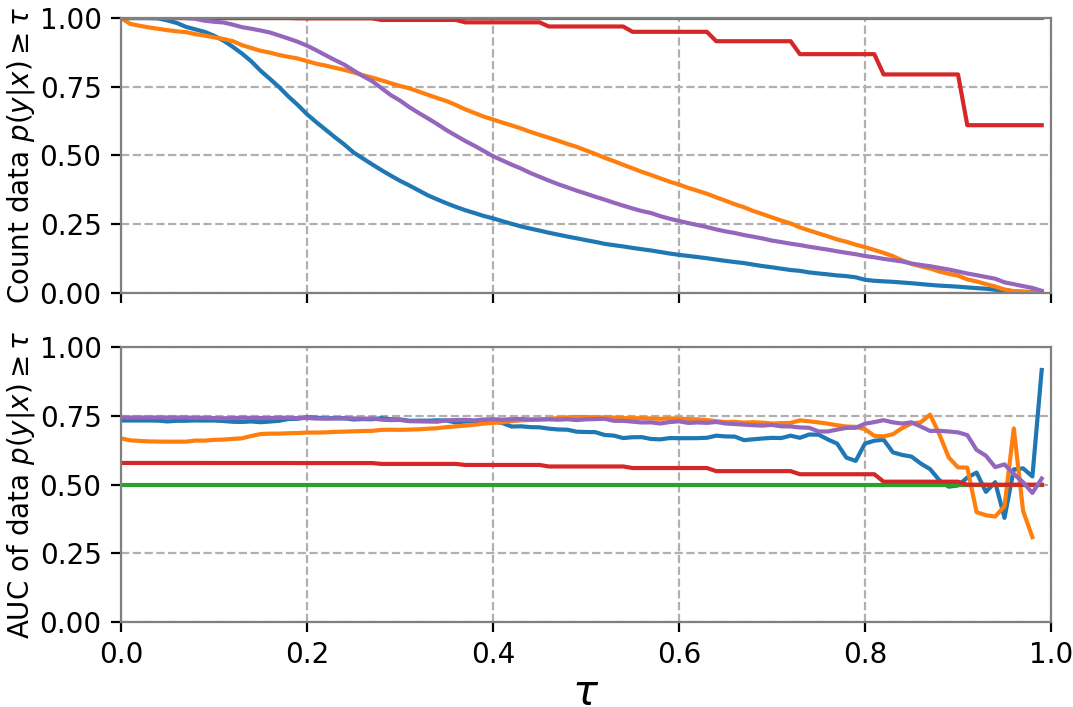}}

\caption{Results on Code Completion Task: \ref{cc_time1} and \ref{cc_time2} show AUC and Brier score as the data is increasingly shifted (across timeline). \ref{cc_author} also presents the two scores as the data is increasingly shifted (across author). We also explore the filtered Number of samples (Count) and AUC versus confidence threshold on the shifted data in \ref{cc_project1}, \ref{cc_project2} and \ref{cc_project3}. In the three shifted project datasets, the number of samples decreases as the confidence score grows. \textit{MC-Dropout} has lower AUC on three shift scenarios since all of its confidence scores are in a very small range around 1. \textit{Vanilla} and \textit{Temp Scale} have the overall decent AUC and Brier. Their effectiveness is stable as the intensity of shift increases.}
\label{cc1}
\end{figure*}

%% file: Table/Uncertainty_ood_time.tex
\begin{table}[h]

\caption{Uncertainty approach evaluation on different timelines (in-/out-of-distribution detection)}
\resizebox{0.49\textwidth}{!}{\begin{tabular}{c|c|ccc|ccc}

\toprule
Dataset & \multirow{2}{*}{Approach} & \multicolumn{3}{c|}{CS} & \multicolumn{3}{c}{CC} \\

In/OOD & & AUC & AUPR & Brier & AUC & AUPR & Brier \\

\midrule
\multirow{5}{*}{val/test1} & Vanilla & 49.04 & 48.38 & 37.60 & 49.67 & 52.43 & 34.15 \\
& Temp Scale & 49.04 & 48.38 & \textbf{37.52} & 49.72 & 52.40 & \textbf{32.74} \\
& mMutate & 50.51 & \textbf{65.39} & 44.08 & \textbf{54.98} & 68.29 & 40.32 \\
& MC-Dropout & \textbf{50.97} & 54.92 & 39.84 & 51.07 & \textbf{76.37} & 47.26 \\
& Dissector & 49.75 & 48.82 & 41.38 & 49.48 & 51.92 & 33.31 \\

\midrule
\multirow{5}{*}{val/test2} & Vanilla  & 50.13 & 49.51 & 37.47 & 47.35 & 48.59 & 35.03 \\
& Temp Scale  & 50.14 & 49.52 & \textbf{37.36} 
& 47.79 & 48.79 & 34.02 \\
& mMutate & 50.67 & \textbf{65.80} & 43.63 
& 50.82 & 65.49 & 43.54 \\
& MC-Dropout & \textbf{51.00} & 55.21 & 39.48 & \textbf{50.88} & \textbf{75.59} & 48.83 \\
& Dissector & 49.71 & 49.50 & 41.51 & 50.06 & 50.34 & \textbf{33.25}\\

\midrule
\multirow{5}{*}{val/test3} & Vanilla & 51.38 & 51.34 & 37.34 & 48.86 & 49.80 & 33.87 \\
& Temp Scale  & 51.38 & 51.34 & \textbf{37.21} & 49.40 & 49.99 & \textbf{32.96} \\
& mMutate & 50.46 & \textbf{66.27} & 43.19 & 43.08 & 61.88 & 46.11 \\
& MC-Dropout & \textbf{51.61} & 56.32 & 38.70 & \textbf{50.95} & \textbf{75.25} & 49.50 \\
& Dissector & 50.85 & 51.40 & 40.57 & 49.74 & 49.64 & 33.33 \\

\bottomrule
\end{tabular}}
\label{ood_eval_time}
\end{table}

%% file: Table/Uncertainty_ood_project.tex
\begin{table}[h]
\caption{Uncertainty approach evaluation on different projects (in-/out-of-distribution detection)}

\resizebox{0.49\textwidth}{!}{\begin{tabular}{c|c|ccc|ccc}

\toprule
Dataset & \multirow{2}{*}{Approach} & \multicolumn{3}{c|}{CS} & \multicolumn{3}{c}{CC} \\
In/OOD & & AUC & AUPR & Brier & AUC & AUPR & Brier \\
\midrule
\multirow{5}{*}{val1/test1} & Vanilla & 50.99 & 60.51 & \textbf{33.70} & 56.42 & 67.99 & 32.24 \\
& Temp Scale & 47.82 & 57.47 & 58.23 & \textbf{57.07} & 68.40 & \textbf{28.33} \\
& mMutate & 50.48 & \textbf{64.23} & 34.86 & 53.95 & 76.95 & 34.46 \\
& MC-Dropout & 54.02 & 60.51 & 38.69 & 50.06 & \textbf{80.88} & 38.24 \\
& Dissector & \textbf{54.28} & 62.14 & 39.83 & 54.16 & 65.26 & 30.03 \\

\midrule
\multirow{5}{*}{val2/test2} & Vanilla & 62.47 & 57.93 & 53.87 & 58.12 & 51.44 & \textbf{26.12}\\
& Temp Scale &  50.96 & \textbf{68.87} & 38.41 &  \textbf{58.52} & 52.04 & 28.17 \\
& mMutate & 61.11 & 62.37 & 42.52 &  57.54 & 63.72 & 49.74 \\
& MC-Dropout & \textbf{67.90} & 61.66 & \textbf{27.09} &  50.08 & \textbf{70.19} & 59.62 \\
& Dissector & 65.73 & 59.75 & 35.66 & 55.09 & 45.66 & 30.59 \\

\midrule
\multirow{5}{*}{val3/test3} & Vanilla &  49.33 & 54.02 & 68.27 & 59.03 & 59.41 & 27.45 \\
& Temp Scale & 50.07 & \textbf{65.05} & 30.08 & 59.38 & 58.57 & \textbf{27.22}  \\
& mMutate & 55.15 & 43.60 & 41.24 & \textbf{59.72} & 71.70 & 45.97 \\
& MC-Dropout & \textbf{58.71} & 37.02 & \textbf{26.60} & 50.02 & \textbf{73.61} & 52.77  \\
& Dissector & 57.86 & 38.25 & 39.95 & 53.18 & 51.83 & 30.46\\

\bottomrule
\end{tabular}}
\label{ood_eval_project}
\end{table}

%% file: Table/Uncertainty_ood_author.tex
\begin{table}[h]
\caption{Uncertainty approach evaluation on different authors (in-/out-of-distribution detection)}
\resizebox{0.49\textwidth}{!}{\begin{tabular}{c|c|ccc|ccc}

\toprule
Dataset & \multirow{2}{*}{Approach} & \multicolumn{3}{c|}{CS} & \multicolumn{3}{c}{CC} \\
In/OOD & & AUC & AUPR & Brier & AUC & AUPR & Brier \\
\midrule
\multirow{5}{*}{val/test1} & Vanilla &  50.79 & 54.93 & \textbf{37.31} & 50.60 & 50.60 & 33.12 \\
& Temp Scale &  50.88 & 55.06 & 39.31 & 50.36 & 50.54 & 32.74\\
& mMutate &  49.74 & \textbf{67.81} & 40.86 & \textbf{59.88} & 70.50 & 41.87 \\
& MC-Dropout &  50.51 & 60.19 & 37.54 & 50.03 & \textbf{75.17} & 49.67\\
& Dissector &  \textbf{50.93} & 55.23 & 38.75 & 50.74 & 50.60 & \textbf{32.03}\\

\midrule
\multirow{5}{*}{val/test2} & Vanilla & \textbf{59.42} & 74.55 & \textbf{30.57} & 51.44 & 60.88 & 34.06 \\
& Temp Scale & 58.71 & 74.21 & 37.06 & 51.95 & 61.14 & 30.79 \\
& mMutate & 55.38 & \textbf{75.81} & 32.29 & 49.85 & 72.11 & 37.15 \\
& MC-Dropout & 59.35 & 74.30 & 30.59 & 50.09 & \textbf{79.59} & 40.82\\
& Dissector & 57.73 & 73.54 & 33.43 & \textbf{52.99} & 60.70 & \textbf{30.05} \\

\midrule
\multirow{5}{*}{val/test3} & Vanilla & \textbf{60.06} & 72.03 & \textbf{31.54} & 52.57 & 63.41 & 33.90 \\
& Temp Scale & 59.49 & 71.68 & 35.13 & 53.08 & 63.78 & 30.04  \\
& mMutate & 53.61 & \textbf{72.41} & 35.67 & \textbf{59.71} & 77.49 & 34.22 \\
& MC-Dropout & 59.20 & 70.94 & 31.70 & 50.11 & \textbf{80.40} & 39.21  \\
& Dissector & 58.51 & 70.99 & 34.11 & 53.19 & 62.80 & \textbf{29.75}\\

\bottomrule
\end{tabular}}
\label{ood_eval_author}
\end{table}

%% file: Threats.tex
\section{Threats to Validity}
\label{sec:threats}

Inappropriate selection of datasets and uncertainty methods may weaken the \textit{external validity} of experimental conclusions. We try to mitigate this threat by the following approaches:
(1) Select sufficient number of train, validation and test files that are versatile with different distributions \cite{hendrycks2019augmix, liang2017enhancing}, 
\eg Datasets for different timelines contain total 36,588 Java files, among which 26,436 files for the \textit{train set}, 2,538 files for the \textit{validation set} and each of the three shifted \textit{test set}. Datasets for different projects contains total 31,115 Java files in three pairs. Specifically, 4,615 files for \textit{train set1}, 1,977 files for \textit{validation set1} and 2,558 files for \textit{test set1}; 5,810 files for \textit{train set2}, 2,489 files for \textit{validation set2} and 2,834 files for \textit{test set2}; 4,684 files for \textit{train set3}, 2,007 files for \textit{validation set3} and 4,141 files for \textit{test set3}.
Datasets for different authors contain total 18,014 Java files, among which 7,137 files for \textit{train set}, 3,059 files for \textit{validation set}, 2,760 files for \textit{test set1}, 2,378 files for \textit{test set2} and 2,680 files for \textit{test set3};
(2) Define three data distribution shift types to manifest the change of model performance; (3) Choose 5 state-of-the-art uncertainty methods with diverse architectures and evaluate their effectiveness in terms of both error/success prediction as well as in/out-of-distribution detection.


Our \textit{internal threat} mainly arises from shifted datasets configuration. Existing uncertainty evaluation in terms of in-/OOD detection on CV or NLP tasks uses complete OOD datasets from different discipline, while in software engineering cases, we focus on program distribution shift and in which the OOD dataset has relatively lower shift intensity. Thus in our experiment, existing uncertainty's effectiveness may get compromised compared to their original experimental results. However, our experiment only focus on the pattern of change of uncertainty effectiveness as the degree of distribution shift increases. In this assumption as long as the uncertainty can more precisely distinguish between in- and OOD samples, it is sensitive to distribution shift.


%% file: Related.tex
\section{Related Work}
\label{sec:related}


\subsection{Predictive Uncertainty for DL Application}
Uncertainty is a natural part of any predictive system, thus modeling uncertainty is of crucial importance.
Existing work has been developed for quantifying predictive uncertainty in DL models and divided into two categories: \textit{Bayesian} methods and \textit{Non-Bayesian} methods.
Popular \textit{Bayesian} approximation approaches include Laplace approximation~\cite{mackay1992bayesian}, variational inference~\cite{blundell2015weight} and dropout-based variational inference \cite{gal2016dropout, kingma2015variational} which activates the dropout layer in the testing phase to measure the uncertainty.
For \textit{Non-Bayesian} methods, \cite{hendrycks2016baseline} first propose a baseline for detecting misclassified samples, which leverages the maximum value of softmax layer as the uncertainty score.
\cite{guo2017calibration} propose re-calibration of probabilities on a held-out validation set through \textit{Temp Scale}.

\subsection{Adversarial Attacks}

Adversarial attack techniques~\cite{kurakin2016adversarial,papernot2016limitations} could generate adversarial perturbations to fool DL models. 
Adversarial perturbations are unnoticeable for human-beings, but they could change the model prediction when applied to normal inputs. 
One well-known example is the Fast Gradient Method (FSGM) search along the gradients' direction to generate adversarial samples.
Besides the gradient search techniques, another type of attack algorithm~\cite{cw,moosavi2016deepfool} models the adversarial sample generation as an optimization problem, which targets to minimize the norm between normal input and adversarial samples that satisfy the constraints.

\subsection{Verification for DL Application}
The following work is proposed to formally verify DL models to ensure their effectiveness. 
\cite{katz2017reluplex, pulina2010abstraction} apply symbolic techniques on the hidden neurons of the DL model to abstract the input space. However, the cost of symbolic techniques that limits this type of techniques can hardly be applied to large DL models.
Some other techniques try to validate inputs at the running time. 
For example, \cite{wang2020dissector} propose a validation technique by measuring the \textit{PV-score} of each input, \cite{wang2019adversarial} mutate the original model to measure the \textit{label change rate (LCR)} to identify the adversarial samples at the running time.

%% file: Future.tex
\section{Future Work}
\label{sec:future}

Although our work, to our best knowledge, is the first to systematically research uncertainty under program distribution shift problem, we mainly focus on Java language. In the future, we would further explore distribution shift across different programming languages. For example, evaluate the performance and uncertainty of Python program classifier on Java snippets. Moreover, we plan to further study program distribution shift problem under more program-analysis tasks such as code authorship identification (AI) \cite{kang2019assessing, abuhamad2018large}, code API search \cite{gu2018deep}, code clone detection \cite{white2016deep}, etc., to make our conclusion more comprehensive and convincing. 
Finally, from our conclusion in Section~\ref{sec:uncertainty_eval}, all existing uncertainty present certain limitations under program distribution shift scenarios. One main reason is that their observation and assumption on CV and NLP datasets may not be adaptive to program data. Future work could try to enhance the performance of uncertainty measurements following this direction.

%% file: Conclusion.tex
\section{Conclusions}
\label{sec:conclusion}


Distribution shift is prevalent in software engineering system which not only degrades the DL model performance but also induce unreliable overconfident predictions. Our comprehensive study illustrates the specific impact of different types of program distributing shift on DL software and the effectiveness of existing state-of-the-art uncertainty methods under program distributing shift. 
We show that the impact of distribution shift on DL models depends on various factors such as the degree of shift, the type of shift, the DL model architecture, etc. and could be mild or severe. 
Furthermore, existing uncertainty methods originally designed for quantifying model uncertainty under CV and NLP tasks, though exhibit reasonable effectiveness under both error/success prediction and in-/OOD detection, all presents certain limitations in program applications.
For example, layer-level uncertainty such as \textit{Dissector} and \textit{MC-Dropout} perform poorly on simple or shallow neural networks, softmax-based uncertainty such as \textit{Vanilla} and \textit{Temp Scale} highly rely on the model performance and are sometimes vulnerable to distribution shift. Further improvement is needed for adapting existing uncertainty methods to software engineering.
